\documentclass[runningheads]{llncs}

\usepackage{eccv}

\usepackage{eccvabbrv}

\usepackage{graphicx}
\usepackage{booktabs}

\usepackage[accsupp]{axessibility}  

\usepackage[pagebackref,breaklinks,colorlinks,citecolor=eccvblue]{hyperref}

\usepackage{orcidlink}

\usepackage{multicol}
\usepackage{multirow}
\usepackage{bbding}

\begin{document}

\title{Rethinking the Efficiency and Effectiveness of Reinforcement Learning for Radiology Report Generation} 

\titlerunning{Rethinking the Reinforcement Learning for R2G}

\author{\setcounter{footnote}{0}
Zilin Lu\inst{1,2}\orcidlink{0000-0003-2437-283X}\thanks{Equal contribution.} \and
Ruifeng Yuan\inst{1,4,5}\protect\footnotemark[1] \and
Weiwei Cao\inst{1,3,4} \and 
Wanxing Chang\inst{1} \and 
Zhongyu Wei\inst{5} \and 
Sinuo Wang\inst{1} \and 
Yong Xia\inst{2} \and 
Ling Zhang\inst{1} \and 
Jianpeng Zhang\inst{1,3,4} \setcounter{footnote}{3}\thanks{Corresponding to Jianpeng Zhang(\email{jianpeng.zhang0@gmail.com}). The work was done during Zilin’s internship at DAMO Academy.} }

\authorrunning{Z.~Lu et al.}

\institute{DAMO Academy, Alibaba Group \and 
Ningbo No.2 Hospital, China \and
Zhejiang University, China \and
Hupan Lab, 310023, China \and
Fudan University, China }

\maketitle

\begin{abstract}
Radiologists highly desire fully automated AI for radiology report generation (R2G), yet existing approaches fall short in clinical utility. Reinforcement learning (RL) holds potential to address these shortcomings, but its adoption in this task remains underexplored. 
In this paper, we revisit RL in terms of data efficiency and optimization effectiveness for R2G tasks. 
First, we explore the impact of data quantity and quality on the performance of RL in medical contexts, revealing that data quality plays a more critical role than quantity. To this end, we propose a diagnostic diversity-based data sampling strategy that enables comparable performance with fewer samples. 
Second, we observe that the majority of tokens in radiology reports are template-like and diagnostically uninformative, whereas the low frequency of clinically critical tokens heightens the risk of being overlooked during optimization. To tackle this, we introduce Diagnostic Token-weighted Policy Optimization (DiTPO), which directly optimizes for clinical accuracy by using a diagnostic F1 score as the reward signal. 
Unlike standard RL approaches that treat all tokens equally, DiTPO explicitly models the varying importance of different tokens through rule- or gradient-based mechanisms to prioritize clinically relevant content. 
Extensive experiments on the MIMIC-CXR, IU-Xray, and CheXpert Plus datasets demonstrate that our framework achieves state-of-the-art (SOTA) performance while requiring substantially fewer training samples in RL. Notably, on MIMIC-CXR, our framework attains an F1 score of 0.516 using only 20\% of the RL training samples. 
  \keywords{Radiology report generation \and Reinforcement learning \and Multimodal large language model}
\end{abstract}

\section{Introduction}
\label{sec:intro}
Automatic radiology report generation (R2G) is a longstanding aspiration of artificial intelligence (AI) in medical imaging. It offers the potential to reduce clinical reporting workload and enhance diagnostic efficiency, especially in resource-constrained settings~\cite{R2G-survey}. Yet, the precise interpretation of complex imaging features and their alignment with domain-specific medical language remains highly demanding, making R2G one of the grand challenges in this field~\cite{tu2024towards, liao2023deep}. 

Multimodal large language models (MLLMs) have advanced rapidly, demonstrating powerful capabilities to integrate and align visual, linguistic, and other modalities~\cite{zhang2024mm,yin2024survey}. These advances have driven significant breakthroughs in multimodal understanding and generation, opening new opportunities for R2G. Recent studies, particularly in the context of chest X-ray imaging, have shown promising progress in applying MLLMs to R2G tasks~\cite{R2Gen,M2TR, PromptMRG, EKAGen, AM-MRG, SS-ACL, MLRG}. 
Most existing MLLMs are trained under the supervised fine-tuning (SFT) paradigm, which typically optimizes the maximum likelihood estimation (MLE) objective to maximize the log-likelihood of each token in the generated sequence. However, this optimization strategy inherently encourages MLLMs to mimic the linguistic style of reference reports by reproducing high-frequency terms and common sentence structures, while at the same time tending to overlook low-frequency but clinically important findings such as \textit{tumor} or \textit{fracture}.
Consequently, the SFT paradigm often leaves MLLMs in a suboptimal state where superficial sentence fluency is prioritized over essential diagnostic content. The resulting reports, though grammatically sound and fluent, may lack sufficient clinical utility.

\begin{figure}[t]
    \centering
    \includegraphics[width=1\linewidth]{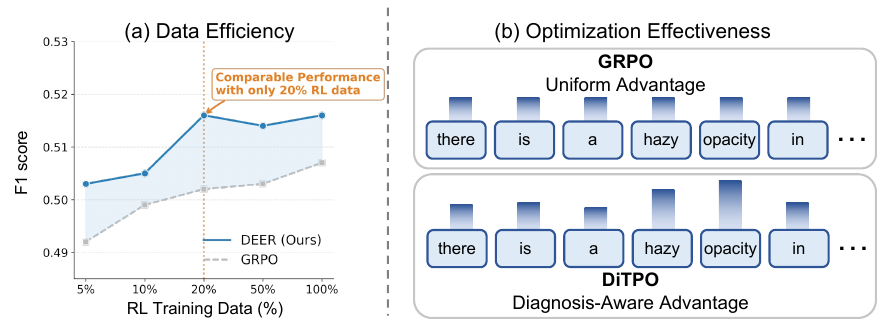}
    \caption{
    (a) Diagnostic F1 scores of GRPO and DEER with different proportions of RL training data. The results indicate that with only about 20\% of the RL training data, the models'performance is comparable to that achieved with 100\% of the data.
    (b) Comparison between uniform GRPO and our DiTPO. GRPO assigns equal advantages to all tokens regardless of their clinical importance, while DiTPO assigns significantly higher advantages to diagnostically critical tokens. 
    }
    \label{fig:1}
\end{figure}

Reinforcement learning (RL) offers a promising approach to overcome the limitations of SFT~\cite{arulkumaran2017deep}. Instead of imitating reference texts, RL directly optimizes clinical objectives by using diagnostic accuracy metrics as rewards. Popular RL-based algorithms, such as Group Relative Policy Optimization (GRPO)~\cite{GRPO}, have been shown to effectively align models with complex reward functions. Preliminary studies investigating RL for R2G have shown its feasibility in enhancing the clinical utility of generated reports~\cite{R2GenRL,LM-RRG,MPO,OISA,EditGRPO}. 
Despite its potential, applying RL to the medical domain presents two major challenges. First, at the data efficiency level, fundamental questions remain about the relative importance of data quantity versus quality in the RL phase, as well as effective strategies for identifying and selecting high-quality RL training data. Second, at the algorithmic effectiveness level, the unique characteristics of radiology reports demand optimization approaches that explicitly emphasize low-frequency but clinically critical tokens. Existing work offers limited investigation into these two crucial aspects. 

In this work, we systematically investigate these challenges and present a \textbf{D}ata-\textbf{E}fficient and Diagnosis-\textbf{E}ffective \textbf{R}einforcement learning (\textbf{DEER}) framework for R2G. 
To explore the data efficiency problem, we first investigate whether the entire training dataset is necessary for effective RL fine-tuning. Our initial experiments on MIMIC-CXR yield a striking discovery: as shown in Figure~\ref{fig:1}(a), fine-tuning with RL on just 20\% of randomly selected data already achieves performance comparable to using the full dataset. This finding reveals substantial data redundancy in the RL stage and raises a crucial question: can we strategically select an even smaller, more informative subset to match or even surpass full-dataset performance?
To address this, we propose a label-free \textbf{D}iagnostic \textbf{D}iversity-based data \textbf{Sampling} (\textbf{DDSampling}), a data selection strategy for RL training. Our approach aims to prioritize training data that exhibit greater diversity and uncertainty in the inference process, thereby providing richer learning signals for RL training to understand the spectrum of clinical presentations. Through this strategy, we demonstrate that only 20\% of carefully selected data can match the performance of training on the entire dataset, significantly improving the data efficiency of RL-based medical report generation.

For effectiveness, we identify a fundamental limitation in GRPO when applied to structured medical text: it assigns a single advantage value to each sequence in group sampling, treating all tokens equally~\cite{GRPO}. However, in medical reports, tokens contribute heterogeneously to diagnostic conclusions. Structural phrases (\eg, ``There is") provide little diagnostic information, while specific findings (\eg., ``opacity," ``effusion") are clinically critical. Ignoring this heterogeneity leads to suboptimal learning, as the model receives diluted and potentially misleading training signals.
To overcome this limitation, we propose assigning a higher token-level advantage to clinically important tokens, thereby guiding the model to prioritize clinically relevant content during the optimization process.
We introduce two mechanisms for token-level advantage assignment:
(a) Rule-based TF-IDF (Term Frequency-Inverse Document Frequency) weighting: Leveraging the structured nature of radiology reports, we employ TF-IDF statistics across the reports in group sampling to identify distinctive tokens in each generated report. By upweighting these tokens, we encourage the model to focus on meaningful clinical descriptions rather than repetitive boilerplate language.
(b) Gradient-based diagnostic weighting: We exploit CheXbert~\cite{smit2020chexbert}, a widely-used clinical classifier for report evaluation, to compute gradient-based importance scores for each token with respect to diagnostic predictions. This approach directly identifies which tokens most influence clinical classification decisions, providing diagnostically-grounded supervision signals.
By integrating these mechanisms into the GRPO framework, we formulate our novel algorithm: \textbf{Di}agnostic \textbf{T}oken-weighted \textbf{P}olicy \textbf{O}ptimization (\textbf{DiTPO}). DiTPO dynamically allocates credit at the token level, compelling the model to focus on generating diagnostically precise content.

We conduct extensive experiments on the MIMIC-CXR, IU-Xray, and CheXpert Plus datasets, demonstrating that our DEER framework achieves state-of-the-art performance on clinical efficacy metrics. Our main contributions are summarized as follows:

\begin{itemize}
\item We introduce DDSampling that curates a small yet highly informative training set for RL training. Our experiments show that training on a small portion of selected data achieves performance comparable to training on the entire dataset.

\item We propose DiTPO, a novel RL algorithm to allocate advantages at the token level from a diagnostic perspective. 
By integrating rule‑ or gradient‑based token weighting strategies, DiTPO addresses the limitation of uniform credit assignment and guides the generation toward diagnostically critical content.

\item Our DEER framework achieves state-of-the-art performance on three public benchmarks, outperforming existing methods on clinical efficacy metrics. In particular, this result is achieved using only 20\% of the training data, underscoring the high efficiency and effectiveness of the framework.

\end{itemize}

\section{Related Work}
\subsection{Radiology Report Generation}
Radiology report generation aims to automatically produce diagnostic narratives from medical images. Initial approaches employed CNN-based encoders paired with LSTM-based decoders for auto-regressive text generation~\cite{cnn+lstm}. 
The introduction of Transformer architectures~\cite{transformer,R2Gen} enabled better modeling of long-range dependencies in R2G. To enhance clinical accuracy, researchers have explored multiple directions: including knowledge-graph integration~\cite{KiUT,PPKED}, multi-task learning~\cite{PromptMRG}, and large-scale cross-modal pre-training ~\cite{clinical-bert,SS-ACL}.
More recently, large language models have been adapted for radiology report generation through instruction tuning~\cite{liu2024bootstrapping, hc-llm}, demonstrating impressive text generation capabilities. However, these methods primarily rely on SFT, which suffers from a fundamental objective misalignment: optimizing for textual similarity often fails to ensure clinical correctness. 
These inherent limitations of the SFT paradigm motivate reinforcement learning approaches that can directly optimize for clinical accuracy metrics such as pathology classification F1-scores.

\subsection{Reinforcement Learning}
Reinforcement learning has emerged as a key training paradigm for MLLM. Recently, GRPO stands out for its ease of use and excellent performance, having demonstrated its effectiveness in general domains such as multimodal reasoning~\cite{ meng2025mm, wang2025vl, zhang2025r1, yuan2025vl}. In R2G, many efforts have been made to explore RL-based methods. 
A major focus has been on reward design: LM-RRG~\cite{LM-RRG} employed a clinical quality metric as a direct reward, while MPO~\cite{MPO} introduced a multi-objective framework to align with diverse user preferences. 
Another direction involves enhancing the training algorithm and strategy. For instance, R2GenRL~\cite{R2GenRL} used RL to refine cross-modal alignment mechanisms. More recently, EditGRPO~\cite{EditGRPO} developed a mixed-policy algorithm to improve training stability, and OISA~\cite{OISA} proposed a self-iterative loop where the model generates its own preference data for alignment.
In this work, we further investigate the application of GRPO in the R2G task, with a focus on enhancing both its efficiency and effectiveness.

\section{Methodology}
\begin{figure}[t]
    \centering
    \includegraphics[width=1\linewidth]{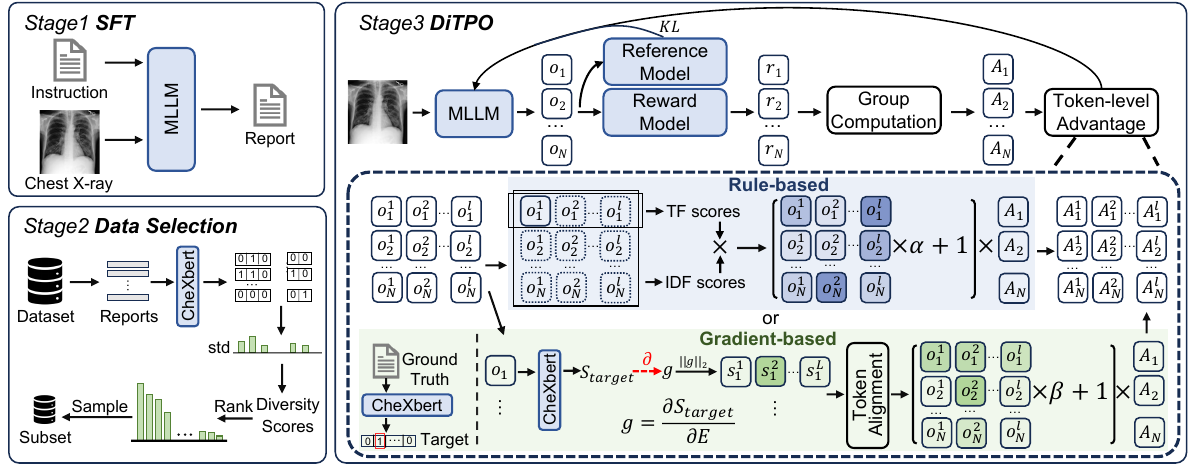}
    \caption{The DEER framework consists of three stages: (1) SFT for cold‑start initialization to provide foundational R2G capabilities; (2) Data Selection via DDSampling to retain high‑quality diverse samples; and (3) DiTPO, where rule‑based (TF‑IDF) and gradient‑based token weighting are integrated to produce diagnosis‑aware token‑level advantages for policy optimization.}
    \label{fig:2}
\end{figure}

The training pipeline of DEER is illustrated in Figure~\ref{fig:2}. In the first stage, SFT is applied as a cold‑start initialization to equip the model with foundational generation capabilities. In the second stage, data selection is performed via DDSampling to identify high‑quality training samples for subsequent learning. In the final stage, DiTPO is employed to optimize the model toward diagnostic objectives.

\subsection{Basic RL Framework}
We build our RL framework upon GRPO. Unlike traditional policy gradient methods that require a separate value network, GRPO estimates advantages using group-level statistics, making it more sample-efficient and stable.
In the context of R2G, given an image-report data point $(I, y)$, GRPO samples a group of $N$ individual reports $\{o_1, o_2, \dots, o_N\}$ from the policy $\pi_\theta$. Each sampled report $o_i$ is evaluated using a reward function $R(o_i, y)$ that measures both clinical diagnostic accuracy and natural language fluency. 
\begin{equation}
r_i = R(o_i, y) = F1(C(o_i), C(y)) + \gamma BLEU(o_i, y),
\label{reward_function}
\end{equation}
where $C$ represents CheXbert~\cite{smit2020chexbert}, a BERT-based classifier trained to extract 14 pathology disease labels (positive or negative) from radiology reports. The $F1$ refers to the computation of the F1-score between predicted labels and ground-truth labels. Meanwhile, $BLEU$ stands for BLEU-2 score that evaluates the natural language similarity between the sampled report and the reference report. 

In practice, the RL training process is structured in two phases to balance the two objectives within the reward function. The initial phase is exclusively dedicated to maximizing the model's clinical diagnostic accuracy, for which the hyperparameter $\gamma$ is set to zero. Subsequently, the second phase aims to refine the model's output fluency without compromising diagnostic performance. Hence, we incorporate the natural language similarity objective as a secondary goal, controlled by a small, non-zero $\gamma$.

After obtaining the reward, the advantage $A_i$ for each report is computed as:

\begin{equation}
A_i = \frac{r_i - \mathrm{mean}(r_1,r_2,\dots,r_N)}{\mathrm{std}(r_1,r_2,\dots,r_N)}.
\label{advantage_GRPO}
\end{equation}
The GRPO objective then applies this advantage estimate within a PPO-style clipped loss function~\cite{ppo}. 
The policy is updated to increase the probability of generating tokens in high-advantage reports while decreasing the probability for low-advantage ones.

\subsection{Diagnostic Diversity-based Data Sampling}
Our preliminary analysis shows that RL training exhibits significant data redundancy. This observation motivates us to design DDSampling that identifies the most informative samples for higher data efficiency in RL optimization.
The effectiveness of GRPO depends on the diversity of responses within each sampled group. Samples where the model generates consistent responses provide minimal learning signals, as advantages within a group are nearly uniform. Conversely, samples with high response variance enable meaningful comparisons and are more valuable for training. For the generation of medical reports, we measure this diversity through uncertainty in clinical predictions across multiple generations from the same input, which we term \textbf{diagnostic diversity}.

For each image-report data point $(I, y)$ in training set, we use the SFT model to generate $K=10$ reports via nucleus sampling. We apply CheXbert to extract binary pathology predictions for each report, obtaining $p_c^{(k)} \in \{0, 1\}$ for each pathology $c$ and sampled report $k$.
For each pathology, we compute the standard deviation across the $K$ reports:
\begin{equation}
\sigma_c = \text{std}(\{p_c^{(1)}, p_c^{(2)}, \dots, p_c^{(K)}\}).
\end{equation}
The overall diagnostic diversity score is:
\begin{equation}
D(I) = \sum_{c=1}^{C} \sigma_c,
\label{diversity_score}
\end{equation}
where $C$ is the number of pathologies. 
We rank all data point by diversity score $D(I)$ in descending order and perform logarithmic rank-based sampling on them. The sampling probability for a data point of rank $r$ from a set of $N$ data points is as follow:

\begin{equation}
P_r = \frac{1}{\log(r + 1) \sum_{j=1}^{N} \frac{1}{\log(j + 1)}}.
\label{sampling_prob}
\end{equation}

This logarithmic weighting prioritizes high-diversity samples while maintaining non-zero probabilities for all data, preventing catastrophic forgetting on well-learned pathology while focusing on diagnostically uncertain cases. During RL training, we sample a data subset according to these probabilities. 

Notably, DDSampling is a label-free method that does not require any ground-truth medical reports or labels. Consequently, this approach can significantly reduce the potential data annotation cost in RL stage by pre-sampling a subset of informative data.

\subsection{DiTPO Algorithm}

In our basic RL framework, a critical observation is that GRPO assigns a single advantage value $A_i$ to the entire report $o_i$. This means all tokens in the report receive the same credit assignment, regardless of their individual contribution to the final reward. While this simplification works well for general text generation, it poses a fundamental limitation for medical reports, where different tokens exhibit vastly different diagnostic importance. 

To address this limitation, we propose DiTPO, which introduces token-level advantages that reflect the heterogeneous diagnostic importance of different tokens in medical reports.
Instead of assigning a single advantage $A_i$ to all tokens in report $o_i$, DiTPO computes a token-specific advantage $A_i^t$ for each token $o_i^t$ at position $t$. 
We decompose the token-level advantage into two components:
\begin{equation}
A_i^t = A_i \cdot w_i^t,
\label{token_advantage}
\end{equation}
where $A_i$ is the report-level advantage from Eq.~\ref{advantage_GRPO}, and $w_i^t$ is a token-specific weight that modulates the importance of token $o_i^t$. By assigning $w_i^t > 1$ to diagnostically important tokens and $w_i^t \approx 1$ to others, we guide the model to prioritize clinically relevant content during optimization.
We present two approaches to compute $w_i^t$: a rule-based method that identifies distinctive clinical expressions, and a gradient-based method that leverages diagnostic model feedback.

\noindent{\textbf{Rule-based token weighting with TF-IDF.}}
Medical reports follow a structured format where certain phrases appear repeatedly across samples (\eg, ``the heart size is" ``no acute" ``stable appearance"). These boilerplate expressions, while grammatically necessary, contribute minimally to diagnostic decisions. Conversely, tokens that uniquely characterize a specific report often correspond to critical findings (\eg, specific pathologies, their locations, or severity descriptors).

To identify such distinctive tokens, we employ the Term Frequency-Inverse Document Frequency (TF-IDF) statistics computed over the group of $N$ sampled reports $\{o_1, o_2, \dots, o_N\}$ for each image. For token $o_i^t$ in report $o_i$, we compute:
\begin{equation}
\text{TF-IDF}(o_i^t) = \text{TF}(o_i^t, o_i) \cdot \text{IDF}(o_i^t, \{o_1, \dots, o_N\}),
\label{tfidf_compu}
\end{equation}
where TF measures the frequency of the token within report $o_i$, and IDF measures its distinctiveness across the group. The rule-based token weight is then defined as:
\begin{equation}
w_i^t = 1 + \alpha \cdot \text{TF-IDF}(o_i^t),
\label{tfidf_weight}
\end{equation}
where $\alpha$ is a hyperparameter controlling the strength of reweighting. This approach encourages the model to focus on unique clinical expressions rather than repetitive template text.

\noindent{\textbf{Gradient-based token weighting with CheXbert.}}
While TF-IDF identifies distinctive tokens based on statistical patterns, it does not explicitly measure their diagnostic relevance. To directly capture which tokens are most critical for accurate clinical predictions, we leverage CheXbert and adopt a gradient-based approach to measure the token importance.

For each training data point $(I, y)$, we identify ground-truth positive diseases set $\mathcal{T}$ from the reference report $y$ using CheXbert. Our model is expected to generate clinically relevant findings of these diseases. 
For a sampled report $o_i$, we aim to identify which tokens most influence CheXbert's predictions for the positive diseases in $\mathcal{T}$. We compute this through gradient-based saliency analysis. 
First, we tokenize the sampled report using CheXbert's tokenizer and obtain input embeddings $\mathrm{E} = [\mathrm{e}_1, \mathrm{e}_2, \dots, \mathrm{e}_L]$, where $L$ is the sequence length in CheXbert's tokenization. These embeddings are set to require gradients. We then perform a forward pass through CheXbert's BERT encoder to obtain the [CLS] representation, which is fed into $C=14$ separate classification heads, one for each disease.
For the positive diseases identified in $\mathcal{T}$, we aggregate their positive and uncertain class logits as our optimization target:
\begin{equation}
S_{\text{target}} = \sum_{c \in \mathcal{T}} \text{logit}_c^{(1)},
\end{equation}
where $\text{logit}_c^{(1)}$ denotes the logit for positive label of the pathology $c$. By summing the logits for positive and uncertain classes across all positive diseases, we create a scalar objective that reflects the model's confidence in predicting the ground-truth pathologies.
We then compute the gradient of $S_{\text{target}}$ with respect to the input embeddings:
\begin{equation}
\mathrm{g} = \frac{\partial S_{\text{target}}}{\partial \mathrm{E}}.
\end{equation}
The gradient magnitude for each CheXbert token in sampled report $i$ is computed as:
\begin{equation}
s_i^j = \|\mathrm{g}_i^j\|_2, \quad j \in [1, L].
\end{equation}
A larger gradient magnitude indicates that the corresponding token has a stronger influence on predicting the target diseases.

Since CheXbert uses a different tokenizer than our report generation model, we need to align the gradient-based importance scores $\{s_i^j\}$ to our model's generated token sequence $\{o_i^t\}$. We perform a token-level mapping: for each token $o_i^t$ in the generated sequence, we aggregate the importance scores from all source CheXbert tokens that overlap with it (\eg, taking the maximum).
%
Let $\tilde{s}_i^t$ denote the aligned importance score for token $o_i^t$ after this mapping. The gradient-based token weight is then:
\begin{equation}
w_i^t = 1 + \beta \tilde{s}_i^t,
\label{gradient_weight}
\end{equation}
where we directly add the importance score to the base weight of 1 with a hyperparameter $\beta$. We do not apply additional normalization, as empirical observations show that the gradient magnitudes naturally exhibit a smooth distribution suitable for direct weighting.

This gradient-based approach provides supervision signals that are directly aligned with diagnostic objectives: tokens that strongly influence CheXbert's predictions for ground-truth diseases receive higher weights, encouraging the model to prioritize generating diagnostically related content. By focusing on ground-truth target diseases, we guide the model to learn which expressions are most effective for conveying specific clinical findings.

\section{Experiments}
\subsection{Experimental Settings}
\noindent{\textbf{Datasets.}} MIMIC-CXR is the largest publicly available dataset for chest X-ray report generation, containing 377,110 images from 227,835 radiological studies. We follow the preprocessing and data split from~\cite{R2Gen}, which split dataset to 270,790/2,130/3,858 for training/validation/test.
CheXpert Plus is a large-scale dataset linking chest X-ray images to de-identified radiology reports, encompassing approximately 223,000 images across 187,700 studies from 64,700 patients.  We follow the data split from MambaXray-VL~\cite{MambaXray-VL}.
IU-Xray consists of 7,470 chest X-ray images paired with 3,955 diagnostic reports. Following PromptMRG~\cite{PromptMRG}, we use the model trained on MIMIC-CXR to perform zero-shot evaluation on IU-Xray dataset, assessing the cross-dataset generalization capability of our approach.

\noindent{\textbf{Evaluation Metrics.}} We evaluate model performance using two complementary types of metrics: natural language generation (NLG) metrics, clinical efficacy (CE) metrics and model metrics. To measure the lexical similarity between generated and reference reports, we employ BLEU-4 (B-4), METEOR (MTR), and ROUGE-L (R-L).
To directly assess diagnostic accuracy, we use CheXbert to extract labels from both generated and ground-truth reports. We then compute precision (P), recall (R), and F1-score (F1) for pathology classification. Specifically, we follow the setting of PromptMRG to compute sample-wise F1 on MIMIC-CXR and IU-Xray, and adopt the setting from MambaXray-VL to compute macro F1 on CheXpert Plus.
Additionally, we report GREEN (GN)~\cite{GREEN} and F1 RadGraph(RG)~\cite{radgraph} to further evaluate clinical entity and relation generation. 

\noindent{\textbf{Implementation Details.}} We build on Qwen2.5-VL-3B~\cite{qwen25-vl} and train in two stages. In SFT stage, we freeze the language model and update only the vision encoder and the vision–language projector, using a learning rate of 1e-5 with a batch size of 64. In RL stage, we unfreeze and optimize the entire model with a learning rate of 1e-6 and a batch size of 64; each update uses 8 rollouts. 
The RL training follows a two-phase reward strategy: initially, we set $\gamma=0$ to focus on clinical accuracy, followed by a refinement stage with $\gamma=0.25$ to enhance linguistic fluency.
For DDSampling, we generate $K=10$ reports per image to compute diagnostic diversity.
We set rule-based DiTPO and gradient-based DiTPO with weights $\alpha$=4 and $\beta$=1, respectively. Based on the experiment result, we adopt gradient-based DiTPO in DEER.

\begin{table}[t]
\caption{Comparison with SOTA methods on the MIMIC-CXR dataset. Results are grouped by paradigm (SFT vs. SFT+RL). "20\%" and "100\%" indicate the percentage of training samples used during the RL stage. \({\flat}\) signifies results from reproduced codes. Best and second-best values are in \textbf{bold} and \underline{underlined}. The GRPO approach in the table refer to our basic GRPO framework described in Section 3.1. Notably, DiTPO refer to DEER framework without data selection stage, and both DiTPO and DEER in main results section use gradient-based weighting.}
\label{table:mimic_results}
\centering
\setlength{\tabcolsep}{1.1mm}
\renewcommand{\arraystretch}{1.1}
\resizebox{\textwidth}{!}{%
\begin{tabular}{c|c|c|ccc|ccc|cc} 
\toprule
\multirow{2}{*}{\textbf{Paradigm}} & 
\multirow{2}{*}{\textbf{Method}} & 
\multirow{2}{*}{\textbf{RL Data}} &
\multicolumn{3}{c|}{\textbf{CE Metrics} $\uparrow$} & 
\multicolumn{3}{c|}{\textbf{NLG Metrics} $\uparrow$} &
\multicolumn{2}{c}{\textbf{Model Metrics} $\uparrow$} \\ 

\cmidrule(lr){4-6}\cmidrule(lr){7-9}\cmidrule(lr){10-11}

& & & \textbf{P} & \textbf{R} & \textbf{F1} & \textbf{B-4} & \textbf{MTR} & \textbf{R-L} & \textbf{GR} & \textbf{RG} \\ 
\midrule

\multirow{11}{*}{SFT} & R2Gen~\cite{R2Gen} & - & 0.333 & 0.273 & 0.276 & 0.103 & 0.142 & 0.277 & - & - \\
& CvT2Dis.~\cite{CvT2Dis} & - & 0.360 & 0.412 & 0.384 & 0.124 & 0.153 & 0.285 & - & - \\
& DCL~\cite{DCL} & - & 0.471 & 0.352 & 0.373 & 0.109 & 0.150 & 0.284 & - & - \\
& R2-LLM & - & 0.465 & 0.482 & 0.473 & 0.128 & \underline{0.175} & 0.291 & - & - \\ 
& RGRG$^{\flat}$~\cite{RGRG} & - & 0.461 & 0.475 & 0.447 & 0.126 & 0.168 & 0.264 & 0.268 & 0.224 \\
& PromptMRG$^{\flat}$~\cite{PromptMRG} & - & 0.501 & 0.509 & 0.476 & 0.112 & 0.157 & 0.268 & 0.264 & 0.228 \\ 
& EKAGen$^{\flat}$~\cite{EKAGen} & - & 0.473 & 0.388 & 0.397 & 0.118 & 0.156 & 0.286 & 0.282 & 0.246 \\
& CoD~\cite{CoD} & - & 0.487 & 0.521 & 0.479 & 0.129 & - & 0.286 & - & - \\ 
& AM-MRG~\cite{AM-MRG} & - & \underline{0.555} & 0.429 & 0.484 & 0.136 & 0.174 & 0.291 & - & - \\ 
& SS-ACL~\cite{SS-ACL} & - & \textbf{0.603} & 0.468 & 0.505 & \textbf{0.144} & \textbf{0.180} & \textbf{0.319} & - & 0.225 \\ 
\cmidrule{1-11}
\multirow{9}{*}{SFT+RL} & R2GenRL~\cite{R2GenRL} & 100\% & 0.342 & 0.294 & 0.292 & 0.109 & 0.151 & 0.287 & - & - \\
& LM-RRG~\cite{LM-RRG} & 100\% & 0.500 & 0.500 & 0.484 & 0.122 & 0.165 & 0.296 & - & - \\
& MPO~\cite{MPO} & 100\% & 0.436 & 0.376 & 0.353 & \underline{0.139} & 0.162 & \underline{0.309} & - & - \\
& OISA~\cite{OISA} & 100\% & - & - & 0.504 & 0.121 & - & - & \textbf{0.327} & 0.254 \\
\cmidrule{2-11}
& GRPO~\cite{GRPO} & 100\% & 0.503 & \underline{0.576} & \underline{0.507} & 0.074 & 0.131 & 0.249 & 0.261 & 0.246 \\
& \textbf{DiTPO} & 100\% & 0.519 & \underline{0.576} & \textbf{0.516} & 0.125 & 0.164 & 0.285 & \underline{0.311} & \textbf{0.289} \\ 
\cmidrule{2-11}
& GRPO~\cite{GRPO} & 20\% & 0.520 & 0.549 & 0.502 & 0.089 & 0.139 & 0.263 & 0.278 & \underline{0.262} \\
& \textbf{DEER} & 20\% & 0.493 & \textbf{0.614} & \textbf{0.516} & 0.097 & 0.141 & 0.253 & 0.258 & 0.231 \\ 
\bottomrule
\end{tabular}%
}
\end{table}

\subsection{Main Results}
We evaluate our proposed framework on three widely used benchmarks: MIMIC-CXR, CheXpert Plus, and IU-Xray. We compare DEER against a comprehensive set of state-of-the-art (SOTA) methods, including both SFT and RL paradigms.

\noindent{\textbf{Performance on MIMIC-CXR.}} 
Table \ref{table:mimic_results} details the comparative results on the MIMIC-CXR. Our DiTPO model with full-data establishes a new SOTA in clinical accuracy, achieving a F1 score of 0.516. This performance surpasses our basic GRPO framework (0.507) and previous leading RL methods like OISA (0.504), as well as strong SFT baselines such as SS-ACL (0.505), validating the effectiveness of directly optimizing for diagnostic entities.
The most compelling finding of our work lies in its exceptional data efficiency. Remarkably, our DEER framework trained with only 20\% of the RL samples achieves an identical $F1$ of 0.516, matching the performance of the model trained on the full dataset. This result is a powerful testament to our core hypothesis: data quality holds greater importance than quantity for RL-based R2G. Our clinical diversity-driven sampling strategy successfully curates a compact yet highly informative data subset, enabling the model to reach peak clinical accuracy while reducing the RL training data requirement by 80\%. This suggests that a substantial portion of the full dataset is redundant for improving clinical accuracy via RL fine-tuning.
We observe that the NLG scores (e.g., BLEU-4, ROUGE-L) of DEER are lower than those model with 100\% data. This is an expected outcome: the full-data model is exposed to more linguistic variations and template-like phrases, which boosts its performance on surface-level text similarity metrics. However, the fact that this broader exposure yields no further gains in clinical accuracy reinforces our argument that prioritizing diagnostic content is more effective and efficient than simply mimicking the syntax of reference reports.

\begin{table}[t]
    \centering
    \begin{minipage}[t]{0.48\textwidth}
        \caption{Comparison with SOTA methods on the CheXpert Plus dataset. Best and second-best values are in \textbf{bold} and \underline{underlined}.}
        \label{table:third_dataset_results}
        \centering
        \renewcommand{\arraystretch}{1.1}
        \resizebox{\linewidth}{!}{%
        \begin{tabular}{c|ccc|ccc} 
        \toprule
        \multirow{2}{*}{\textbf{Method}} & \multicolumn{3}{c|}{\textbf{CE Metrics} $\uparrow$} & \multicolumn{3}{c}{\textbf{NLG Metrics} $\uparrow$} \\ 
        \cmidrule(lr){2-4}\cmidrule(lr){5-7}
        & \textbf{P} & \textbf{R} & \textbf{F1} & \textbf{B-4} & \textbf{MTR} & \textbf{R-L} \\ 
        \midrule
        R2Gen~\cite{R2Gen} & 0.318 & 0.200 & 0.181 & 0.081 & 0.113 & 0.246 \\
        R2GenCMN~\cite{R2GenCMN} & 0.329 & 0.241 & 0.231 & 0.087 & 0.127 & 0.256 \\
        WCL~\cite{WCL} & 0.335 & 0.259 & 0.256 & 0.084 & 0.126 & 0.253 \\
        R2GenRL~\cite{R2GenRL} & 0.193 & 0.229 & 0.196 & 0.035 & 0.101 & 0.186 \\
        XProNet~\cite{XProNet} & 0.314 & 0.247 & 0.259 & 0.100 & 0.146 & 0.265 \\
        ORGan~\cite{ORGan} & 0.288 & 0.287 & 0.277 & 0.086 & 0.135 & 0.261 \\
        CvT2Dis.~\cite{CvT2Dis} & 0.285 & 0.252 & 0.246 & 0.067 & 0.118 & 0.238 \\
        CAMANet~\cite{CAMANet} & 0.328 & 0.224 & 0.216 & 0.083 & 0.118 & 0.249 \\
        Token-Mixer~\cite{Token-Mixer} & 0.309 & 0.270 & 0.288 & 0.091 & 0.135 & 0.270 \\
        R2GenGPT~\cite{R2GenGPT} & 0.315 & 0.244 & 0.260 & 0.101 & 0.145 & 0.266 \\
        PromptMRG~\cite{PromptMRG} & 0.258 & 0.265 & 0.281 & 0.095 & 0.121 & 0.222 \\
        R2GenCSR~\cite{R2GenCSR} & 0.315 & 0.247 & 0.259 & 0.100 & 0.146 & 0.265 \\
        MambaX-VL~\cite{MambaXray-VL} & \underline{0.377} & \underline{0.319} & 0.335 & \textbf{0.112} & \underline{0.157} & \underline{0.276} \\
        AM-MRG~\cite{AM-MRG} & \textbf{0.396} & 0.318 & \underline{0.336} & \underline{0.109} & \textbf{0.173} & \textbf{0.282} \\
        \midrule
        \textbf{DEER} & 0.371 & \textbf{0.387} & \textbf{0.355} & 0.058 & 0.112 & 0.222 \\
        \bottomrule
        \end{tabular}%
        }
    \end{minipage}\hfill 
    \begin{minipage}[t]{0.48\textwidth}
        \caption{Zero-shot generalization performance on the IU-Xray dataset. All methods listed are trained exclusively on MIMIC-CXR and evaluated directly on IU-Xray without any fine-tuning. Best and second-best values are in \textbf{bold} and \underline{underlined}.}
        \label{table:iuxray_results}
        \centering
        \renewcommand{\arraystretch}{1.1}
        \resizebox{\linewidth}{!}{%
        \begin{tabular}{c|ccc|ccc} 
        \toprule
        \multirow{2}{*}{\textbf{Method}} & 
        \multicolumn{3}{c|}{\textbf{CE Metrics} $\uparrow$} & 
        \multicolumn{3}{c}{\textbf{NLG Metrics} $\uparrow$} \\ 
        \cmidrule(lr){2-4}\cmidrule(lr){5-7}
        & \textbf{P} & \textbf{R} & \textbf{F1} & \textbf{B-4} & \textbf{MTR} & \textbf{R-L} \\ 
        \midrule
        R2Gen~\cite{R2Gen} & 0.141 & 0.136 & 0.136 & 0.059 & 0.131 & 0.253 \\
        CvT2Dis.~\cite{CvT2Dis} & 0.174 & 0.172 & 0.168 & 0.082 & 0.147 & 0.277 \\
        M2KT~\cite{M2KT} & 0.153 & 0.145 & 0.145 & 0.078 & 0.153 & 0.261 \\
        DCL~\cite{DCL} & 0.168 & 0.167 & 0.162 & 0.074 & 0.152 & 0.267 \\
        RGRG~\cite{RGRG} & 0.183 & 0.187 & 0.180 & 0.063 & 0.146 & 0.180 \\
        PromptMRG~\cite{PromptMRG} & 0.213 & 0.229 & 0.211 & 0.098 & \underline{0.160} & 0.281 \\ 
        CoD~\cite{CoD} & \underline{0.218} & \underline{0.234} & 0.219 & 0.091 & - & \underline{0.288} \\ 
        OISA~\cite{OISA} & - & - & \underline{0.225} & \textbf{0.122} & - & - \\
        \midrule
        \textbf{DEER} & \textbf{0.228} & \textbf{0.261} & \textbf{0.230} & \underline{0.119} & \textbf{0.176} & \textbf{0.324} \\ 
        \bottomrule
        \end{tabular}%
        }
    \end{minipage}
\end{table}

\noindent{\textbf{Performance on CheXpert Plus.}} 
As presented in Table \ref{table:third_dataset_results}, DEER again demonstrates superior clinical performance. It achieves the highest clinical F1 score ($F1$) of 0.355, outperforming strong competitors like AM-MRG (0.336). Similarly, DEER is relatively low for the NLG metrics.

\noindent{\textbf{Zero-Shot Generalization on IU-Xray.}} 
To verify the model's generalization ability, we directly transferred the models trained on MIMIC-CXR to the IU-Xray for evaluation in a zero-shot setting. As presented in Table \ref{table:iuxray_results}, DEER demonstrates the best diagnostic performance, achieving a state-of-the-art clinical F1 score of 0.230. This superior zero-shot performance highlights its ability to learn transferable clinical knowledge rather than overfitting to the source dataset's specific reporting style. Furthermore, DEER also excels in linguistic quality, achieving the best METEOR score (0.176) and a highly competitive ROUGE-L score (0.324). Here, we observe a performance gap on NLG metrics: DEER underperforms SFT-based approaches on in-domain experiments, but it almost surpasses them on zero-shot experiment. This suggest the SFT methods tend to overfit the language distribution on in-domain data, while our RL-based framework shows it robustness in zero-shot setting.

\begin{table}[tb]
    \centering
    \begin{minipage}[t]{0.48\textwidth}
        \caption{Ablation study on the MIMIC-CXR dataset to validate the effectiveness of our proposed components.}
        \label{table:ablation}
        \centering
        \renewcommand{\arraystretch}{1.1}
        \resizebox{\linewidth}{!}{%
        \begin{tabular}{c|c|cccc} 
        \toprule
        \textbf{Method} & \textbf{Weighting} &  \textbf{P} & \textbf{R} & \textbf{F1} & \textbf{B-4} \\ 
        \midrule
        SFT & - & 0.441 & 0.386 & 0.385 & 0.094 \\
        DiTPO & Rule & 0.525 & 0.559 & 0.511 & 0.124\\
        DiTPO & Grad & 0.519 & 0.576 & \textbf{0.516} & \textbf{0.125}\\
        DEER & Rule & \textbf{0.535} & 0.534 & 0.504 & 0.099\\
        DEER & Grad & 0.493 & \textbf{0.614} & \textbf{0.516} & 0.097\\
        \bottomrule
        \end{tabular}%
        }
    \end{minipage}\hfill
    \begin{minipage}[t]{0.48\textwidth}
        \caption{Analysis of reward strategies for optimal balance between clinical accuracy and linguistic quality on DEER.}
        \label{table:reward}
        \centering
        \renewcommand{\arraystretch}{1.1} 
        \resizebox{\linewidth}{!}{%
        \begin{tabular}{c|c|cccc} 
        \toprule
        \textbf{Reward} & $\gamma$ & \textbf{P} & \textbf{R} & \textbf{F1} & \textbf{B-4} \\ 
        \midrule
        F1 only & - & 0.486 & \textbf{0.617} & 0.514 & 0.046 \\
        \midrule
        F1+BLEU-2 & 0.25 & 0.481 & 0.615 & 0.510 & 0.081 \\
        \midrule
        F1 $\rightarrow$ & \multirow{2}{*}{0.25} & \multirow{2}{*}{\textbf{0.493}} & \multirow{2}{*}{0.614} & \multirow{2}{*}{\textbf{0.516}} & \multirow{2}{*}{\textbf{0.097}}\\
        F1 + BLEU-2 & & & & & \\
        \bottomrule
        \end{tabular}%
        }
    \end{minipage}
\end{table}

\subsection{Ablation Study}
We conduct an ablation study on the MIMIC-CXR dataset to analyze the effectiveness of our proposed components. The results, presented in Table \ref{table:ablation}, offer several key insights.
First and foremost, the results clearly validate the benefit of employing reinforcement learning over a standard SFT baseline. All our RL-based variants demonstrate a substantial leap in diagnostic accuracy.
Second, comparing this two token weighting strategies under the same data setting, the gradient-based approach consistently outperforms rule-based approach. This suggests that leveraging direct feedback from the CheXbert classifier via gradients provides more accurate identification of diagnostically critical tokens than statistical TF-IDF measures alone.
Thirdly, with the application of the DDSampling strategy, a key distinction emerged under data-scarce conditions. Gradient-based methods continued to exhibit robust diagnostic capabilities, even on a reduced dataset. Conversely, the performance of rule-based methods degraded, which can likely be attributed to the lower dependency of gradient-based approaches on the size of the dataset.

\subsection{Analysis of Reward Functions}
The design of the reward function is crucial in RL as it defines the model's learning objective. We studied how to best balance clinical accuracy with text quality in our reward signal. As shown in the Table~\ref{table:reward}, using only the F1 score as the reward gives the best clinical performance (F1 of 0.514) but results in a low BLEU-4 score (0.046). We tried directly mixing F1 and BLEU-2 rewards from the start. This approach consistently hurt clinical performance. For example, while it raised the BLEU-4 score to 0.081, the clinical F1 dropped to 0.510. This shows that a single, mixed reward creates a conflict, forcing the model to sacrifice accuracy for linguistic style. 
Based on this insight, we adopted a two-phase reward strategy. In the first phase, the model is trained with a F1-only reward to establish a strong foundation in clinical accuracy. In the second phase, a BLEU-2 reward is introduced to refine the model's linguistic style. This approach proved highly effective: as shown in Table \ref{table:reward}, it maintains the peak clinical F1 score of 0.516 while significantly boosting the BLEU-4 to 0.097.

\begin{table}[tb]
    \centering
    \begin{minipage}[t]{0.48\textwidth}
        \caption{Validation of token weighting via masking. }
        \label{table:token_masking}
        \centering
        \renewcommand{\arraystretch}{1.2} 
        \resizebox{\linewidth}{!}{%
        \begin{tabular}{c|cc} 
        \toprule
        \textbf{Masking Strategy} & \textbf{Label Mod. Ratio} $\uparrow$ & \textbf{Post-masking F1} $\downarrow$ \\ 
        \midrule
        Random & 16.4\% & 0.93 \\
        TF-IDF (Rule) & 25.6\% & 0.89 \\
        Gradient (Ours) & \textbf{32.7\%} & \textbf{0.83} \\
        \bottomrule
        \end{tabular}%
        }
    \end{minipage}\hfill
    \begin{minipage}[t]{0.48\textwidth}
        \caption{Analysis of reward diversity to evaluate the effectiveness of the DDSampling strategy.}
        \label{table:reward_diversity}
        \centering
        \renewcommand{\arraystretch}{1.2} 
        \resizebox{\linewidth}{!}{%
        \begin{tabular}{c|ccc} 
        \toprule
        \textbf{Data Setting} & \textbf{Zero-variance} $\downarrow$ & \textbf{Unique Rewards} $\uparrow$ & \textbf{Avg. Std} $\uparrow$ \\ 
        \midrule
        Random (100\%) & 35.5\% & 2.6 & 0.15 \\
        DDSampling (20\%) & \textbf{16.9\%} & \textbf{3.6} & \textbf{0.19} \\
        \bottomrule
        \end{tabular}%
        }
    \end{minipage}
\end{table}

\subsection{Validation of Token Weighting via Masking} To empirically validate that our token weighting strategies effectively identify diagnostically critical tokens, we conducted a Token Masking experiment on 10,000 random rollout samples. As shown in Table \ref{table:token_masking}, we masked the top 10\% of tokens identified as "important" by each strategy and measured the diagnostic impact. The gradient-based method achieved a Label Modification Ratio (diagnostic change rate) of 32.7\%, significantly outperforming the TF-IDF (25.6\%) and Random (16.4\%) baselines. Furthermore, the post-masking F1 score dropped most severely under gradient-based masking (0.83) compared to TF-IDF (0.89) and Random (0.93). These results provide strong evidence that the gradient-based approach accurately isolates and prioritizes the tokens most responsible for clinical correctness, justifying its selection as the default configuration in the DEER framework.

\subsection{Analysis of DDSampling and Reward Diversity}
A core premise of our DDSampling strategy is that semantic diversity during the sampling stage yields more diverse advantages and informative learning signals for the RL optimization process. To verify this, we analyzed the reward signals generated from the full dataset compared to our selected 20\% subset. As detailed in Table \ref{table:reward_diversity}, our analysis reveals that DDSampling drastically reduces the ratio of zero-variance groups (where all samples receive a reward of 0 or 1) from 35.5\% to 16.9\%. Furthermore, within the selected subset, the number of unique reward values per group rises significantly from 2.6 to 3.6, providing a much more fine-grained supervision signal. Consequently, the average reward standard deviation increases by 26.7\% ($0.15 \rightarrow 0.19$). This confirms that prioritizing training data with high diagnostic uncertainty effectively mitigates data redundancy and maximizes the efficiency of the RL stage.

\begin{figure}[t]
    \centering
    \includegraphics[width=1\linewidth]{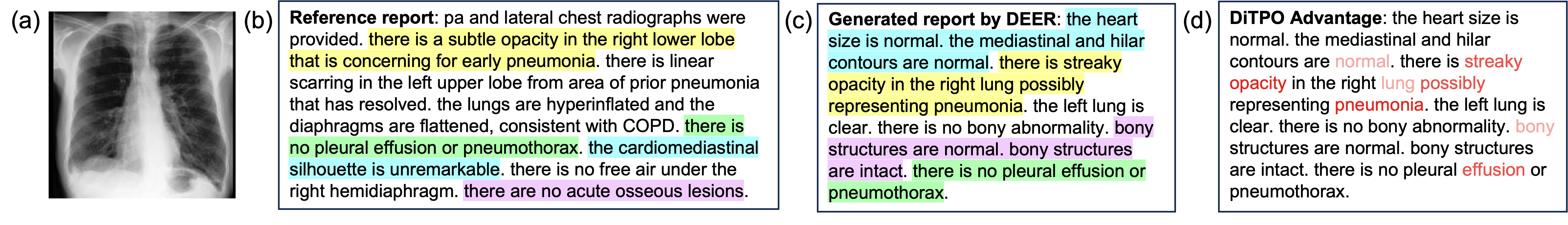}
    \caption{Diagnostic token weighting case study. Sentences describing the same clinical finding in both the reference and generated reports are highlighted with matching colors. The generated report uses red color intensity to visualize token-level weights.}
    \label{fig:3}
\end{figure}

\subsection{Case Study}
Figure~\ref{fig:3} presents a representative case study demonstrating both the clinical accuracy and token weighting effectiveness of our approach. The generated report successfully captures all major pathological findings present in the reference report, such as the pneumonia. The red color intensity visualization reveals that our DiTPO method effectively identifies diagnostically critical tokens. This demonstrates that our token-level optimization focuses learning on clinically relevant content while maintaining comprehensive disease coverage.

\section{Conclusion}
In this paper, we revisit the application of RL for R2G, with a dual focus on enhancing data efficiency and clinical accuracy. We introduced DEER, a novel framework that directly optimizes for clinical correctness by assigning greater importance to diagnostically critical tokens during policy updates. This approach is complemented by a clinical diversity-driven sampling strategy that significantly reduces the need for large-scale RL training data.
Extensive experiments on MIMIC-CXR, CheXpert Plus, and IU-Xray robustly validate the effectiveness of the DEER framework. Our key findings are twofold: First, DiTPO achieves state-of-the-art clinical accuracy, consistently outperforming previous methods across multiple datasets. Second, and more notably, we demonstrate that the DEER framework can achieve this peak performance using only 20\% of the RL training data.



%
%
\bibliographystyle{splncs04}
\bibliography{main}
\end{document}